\documentclass[a4paper,fleqn]{cas-sc}

\usepackage[authoryear,longnamesfirst]{natbib}

\def\tsc#1{\csdef{#1}{\textsc{\lowercase{#1}}\xspace}}
\tsc{WGM}
\tsc{QE}
\tsc{EP}
\tsc{PMS}
\tsc{BEC}
\tsc{DE}
\begin{document}
\let\WriteBookmarks\relax
\def\floatpagepagefraction{1}
\def\textpagefraction{.001}

\shorttitle{Pathway to data-driven geotechnics}

\shortauthors{S Wu et~al.}

\title [mode = title]{Pathway to a fully data-driven geotechnics: lessons from materials informatics}                      
\tnotemark[1]

\tnotetext[1]{This document is the results of the research
   project funded by the KAKENHI.}

%\tnotetext[2]{The second title footnote which is a longer text matter
%   to fill through the whole text width and overflow into
%   another line in the footnotes area of the first page.}

\author[1]{Stephen Wu}[type=editor,
                        auid=000,bioid=1,
                        prefix=Dr,
                        role=Associate professor,
                        orcid=0000-0002-7847-8106]

\cormark[1]

\fnmark[1]

\ead{stewu@ism.ac.jp}

\ead[url]{https://researchmap.jp/stewu?lang=en, http://daweb.ism.ac.jp/~stewu/}

\credit{Conceptualization of this study, Methodology, Writing - draft preparation}

\affiliation[1]{organization={Research Organization of Information and Systems, The Institute of Statistical Mathematics},
    addressline={Midori-cho 10-3}, 
    city={Tachikawa, Tokyo},
    % citysep={}, % Uncomment if no comma needed between city and postcode
    postcode={190-8562}, 
    country={Japan}}

\affiliation[1]{organization={The Graduate University for Advanced Studies, Department of Statistical Science},
    addressline={Midori-cho 10-3}, 
    city={Tachikawa, Tokyo},
    % citysep={}, % Uncomment if no comma needed between city and postcode
    postcode={190-8562}, 
    country={Japan}}

\author[2]{Yu Otake}%[style=chinese]
\credit{Methodology, Writing - draft preparation}

\author[3]{Yosuke Higo}
\credit{Methodology, Writing - draft preparation}

\author[4]{Ikumasa Yoshida}%[style=chinese]
\credit{Methodology, Writing - draft preparation}

\affiliation[2]{organization={Department of Civil Environmental Engineering, Tohoku University},
    addressline={6-6-06, Aramaki Aza Aoba, Aoba-ku}, 
    city={Sendai, Miyagi},
    citysep={}, % Uncomment if no comma needed between city and postcode
    postcode={980-8579},
    country={Japan}}

\affiliation[3]{organization={Department of Urban Management, Graduate School of Engineering, Kyoto University},
    addressline={C1-235 Kyotodaigaku-Katsura, Nishikyo-ku}, 
    city={Kyoto},
    citysep={}, % Uncomment if no comma needed between city and postcode
    postcode={615-8540},
    country={Japan}}

\affiliation[4]{organization={Department of Urban and Civil Engineering, Tokyo City University},
    addressline={1-28-1 Tamazutsumi, Setagaya-ku}, 
    city={Meguro, Tokyo},
    citysep={}, % Uncomment if no comma needed between city and postcode
    postcode={158-8557}, 
    country={Japan}}

\cortext[cor1]{Principal corresponding author}

% % Footnote text
% \fntext[fn1]{This is the first author footnote. but is common to third
% \fntext[fn2]{Another author footnote, this is a very long footnote and
%   it should be a really long footnote. But this footnote is not yet
%   sufficiently long enough to make two lines of footnote text.}

% % For a title note without a number/mark
% \nonumnote{This note has no numbers. In this work we demonstrate $a_b$
%   the formation Y\_1 of a new type of polariton on the interface
%   between a cuprous oxide slab and a polystyrene micro-sphere placed

\begin{abstract}
This paper elucidates the challenges and opportunities inherent in integrating data-driven methodologies into geotechnics, drawing inspiration from the success of materials informatics. Highlighting the intricacies of soil complexity, heterogeneity, and the lack of comprehensive data, the discussion underscores the pressing need for community-driven database initiatives and open science movements. By leveraging the transformative power of deep learning, particularly in feature extraction from high-dimensional data and the potential of transfer learning, we envision a paradigm shift towards a more collaborative and innovative geotechnics field. The paper concludes with a forward-looking stance, emphasizing the revolutionary potential brought about by advanced computational tools like large language models in reshaping geotechnics informatics.
\end{abstract}

% \includegraphics{figs/grabs.pdf}

\begin{highlights}
\item Drawing parallels with materials informatics, this paper underscores the unprecedented potential of integrating deep learning techniques to navigate the intricate complexity and heterogeneity inherent in geotechnics.
\item Emphasizing community-driven initiatives, the research champions the need for open databases and data-sharing platforms in geotechnics to foster collaboration and advance the field.
\item Leveraging state-of-the-art computational tools and methodologies, such as transfer learning, the study envisions a transformative and data-centric future for geotechnics.
\end{highlights}

% Keywords

\maketitle

\section{Introduction}

Recent advancements in machine learning, particularly those driven by deep learning and big data, have undeniably revolutionized the scientific landscape. In fields as varied as mathematics~\citep{Davies:2021aa}, biology~\citep{Jumper:2021aa}, and materials science~\citep{Choudhary:2022aa}, these technologies have propelled innovative breakthroughs, transforming theoretical possibilities into tangible realities. New research fields that incorporate informatics have become the most promising interdisciplinary field, such as bioinformatics, chemoinformatics, materials informatics (MI), etc. The impact of deep learning permeates not only specialized research spheres but also our quotidian life. For instance, personalized medicine leverages these advancements for tailored treatment plans~\citep{Zhang:2019}, while everyday applications such as voice assistants~\citep{MEHRISH2023101869} and recommendation systems~\citep{Mu2018} embed deep learning at their core. The role of modern machine learning is multifaceted. It liberates humans from repetitive tasks, refines and surpasses existing problem-solving methodologies, addresses previously intractable challenges, and most intriguingly, delineates new horizons of exploration and discovery~\citep{PhoonZhang:2023}.

Within the realm of geotechnics, the impact of machine learning is also rising continuously. Buoyed by the successes observed in various scientific domains, the past decade has witnessed burgeoning interest in harnessing these technologies for geotechnical applications~\citep{ZHANG2021327,PHOON2022967}. For instance, predictive models powered by Bayesian neural networks have been explored for soil compressibility and undrained shear strength prediction~\citep{Zhang:2022:BNN}, and regression techniques have been employed for soil liquefaction assessment~\citep{Jas:2023} and slope stability analysis~\citep{Lin:2021aa}. However, the journey has not been without its obstacles. Many of these applications remain constricted by issues such as ``ugly data''~\citep{PCS:2022} (e.g., sparse and noisy), model overfitting, or the lack of interpretability. Concurrently, while geoinformatics -- an interdisciplinary field of data science and earth science --- has been gaining traction~\citep{GI:2019}, the idea of ``geotechnics informatics'' (GtI) remains relatively obscure, often overshadowed in the wider geotechnical community~\citep{2025_34_Elmo,PhoonZhang:2023}.

Looking towards materials science provides an enlightening parallel. In many ways, materials science and geotechnics are closely related, especially considering that soils can be conceptualized as intricate natural composites replete with a mosaic of organic and inorganic constituents. The year 2011 marked a pivotal moment for materials science, with the launch of the Materials Genome Initiative announced by the US government~\citep{MGI2011}. This endeavor, rooted in data science, envisioned halving the time required for material discoveries at that time~\citep{White:2012aa}. This sentiment resonated globally, finding echoes in different regions, such as the NOMAD project of the European Union~\citep{Draxl:2018aa,Sbailo:2022aa}, the MARVEL project of Swtizerland~\citep{MARVEL2014}, and other similar projects initiated in Japan, Korea, and China. A decade later, the effectiveness of using data science to accelerate materials discovery is evident~\citep{Choudhary:2022aa}. MI has matured into a mainstay within materials science research~\citep{Agrawal:2016,Ramprasad:2017aa,Himanen:2019}. But what catalyzed this rapid assimilation of data science within materials science, and why does a parallel trajectory elude geotechnics?

This paper embarks on a journey to address this conundrum. Our objective is to carve out a robust data-driven framework for geotechnics, drawing inspiration from the triumphs of MI. We initiate our exploration by delineating the multifaceted challenges inherent to geotechnics, contrasting them against the unique capabilities that deep learning offers, distinguishing it from its traditional machine learning counterparts. Delving deeper, we scrutinize the hurdles impeding the realization of a holistic GtI paradigm. Harnessing insights gleaned from materials informatics, we then spotlight the potential avenues and opportunities ripe for exploration within GtI. Our ambition is to pave a transformative path: one that seamlessly interweaves time-tested experiential wisdom in geotechnics with the vigor and versatility of a data-driven approach, culminating in a synthesis that promises us a completely new way to approach problems in geotechnics.

\section{Background}

\subsection{Brief introduction to problems in geotechnics}
% Geotechnical engineering is the study of the behavior of soils under the influence of loading forces and soil-water interactions. Understanding the behavior of soils and rocks under various stress conditions is crucial for infrastructure development. Problems in geotechnics include:
% \begin{itemize}
%     \item Foundation engineering - Determining the load-bearing capacity of the ground and designing foundations that can safely bear imposed loads is a fundamental concern.
%     \item Slope stability - Predicting and preventing landslides or failures in both natural and man-made slopes.
%     \item Tunnel engineering - Efficient construction methods, monitoring, ground stability
%     \item Ground improvement - Techniques to enhance the strength, stiffness, and performance of soils in situ.
% \end{itemize}
The field of geotechnical engineering is primarily concerned with controlling the stability of three types of structures: foundation structures, earth structures, and excavation. The ground, comprising a three-phase material of soil, water, and air, exhibits a variety of mechanical behaviors due to the interactions among these phases in response to external actions. These interactions range from those occurring during short-term conditions in construction to long-term post-construction scenarios, and even rare events (e.g, earthquakes and floods). Therefore, careful judgment of the design conditions of the target structure is essential, necessitating the implementation of appropriate ground surveys and soil tests under various scenarios.
\begin{itemize}
    \item Foundation Structures - This involves evaluating the stability of structures that support superstructures, such as shallow foundations and pile foundations. While the primary focus is on estimating the ultimate bearing capacity, estimating deformation amounts to prevent impact on the superstructure can also be a critical design task.
    \item Earth Structures - Focus is primarily on estimating the short-term stability of slopes during construction, including natural slopes, cut earth structures, artificial embankments, and reclaimed land. However, for foundations in clay, it is essential to predict deformation characteristics, such as uneven settling associated with the long-term dissipation of excess pore water pressure. Additionally, these structures are significantly affected by very short-term changes during occasional disasters, such as rainwater infiltration during floods and liquefaction during earthquakes.
    \item Excavation - In recent years, there has been an increase in cases requiring the creation of large underground spaces, such as tunnels and subways. Along with the stability of the excavation face, predicting deformation amounts is also necessary from the perspective of the impact on surrounding structures and environments. Moreover, large-scale excavations often involve extensive groundwater pumping, causing changes in the surrounding groundwater flow. Therefore, estimating the stability of excavation foundations due to the flow of water in the soil becomes crucial.
\end{itemize}

Traditional geotechnical solutions are often empirical, based on in-situ testing and laboratory experiments. However, the inherent variability and heterogeneity of soil make it a challenging material to study. Tracing back through decades of rigorous study and experimentation, the annals of geotechnics are now rich with accumulated data relative to the past. This vast repository encourages the contemporary wave of researchers who are keen to harness the potential of machine learning and data science in decoding the intricate puzzles of the discipline. Yet, the progress of exploiting the latest machine learning framework in geotechnics is slower than the other disciplines, with only a handful of publications delving into the application of deep learning and big data technologies~\citep{PHOON2022967}. This situation is not unfounded, as the complexities embedded within geotechnical challenges present a formidable frontier. We will delve into the challenges and prospects of integrating cutting-edge machine learning techniques into the geotechnical paradigm in the following sections.

\subsection{Brief introduction to modern machine learning}
The evolution of scientific inquiry has seen a shift from the empirical, to the theoretical, to the computational, and now, to the advent of the ``Fourth Paradigm'' of science~\citep{hey2009the}. This latest paradigm, intrinsically data-centric in its approach, champions the idea that the sheer volume, complexity, and dynamism of modern datasets can no longer be comprehensively understood solely through traditional experimental or computational methods. Instead, it necessitates the development of sophisticated algorithms and infrastructures to gather, store, and analyze this data. In this realm, patterns emerge from the data itself, rather than being imposed a priori. The Fourth Paradigm underscores the transformation of data into knowledge, thus redefining our methodologies for scientific discovery in the modern age.

The journey to our current era of data-driven inquiry has been marked by pivotal milestones shaping its trajectory. Many of the foundational mathematical tools employed in the modern machine learning framework were developed before 2000. The potential to train deep neural networks efficiently was successfully demonstrated as early as 2006~\citep{hinton:2006}. On the hardware front, the 21st century heralded the era of ``Big Data,'' buoyed by technological leaps in data storage and computational power. Post-2006, landmark competitions, such as the Netflix Prize~\citep{NetflixPrize:2007} and the ImageNet Large Scale Visual Recognition Challenge~\citep{deng2009imagenet,Russakovsky:2015aa}, catalyzed the evolution and popularization of deep learning models like Convolutional Neural Networks (CNNs)~\citep{venkatesan2018convolutional} and Recurrent Neural Networks (RNNs)~\citep{TEALAB2018334}. The potency of deep learning was vividly showcased when AlphaGo~\citep{Silver:2016aa} defeated a professional human player in Go, a complex board game once deemed insurmountable for computers. Furthermore, the democratization of machine learning tools, through platforms like TensorFlow~\citep{tensorflow2015-whitepaper} and PyTorch~\citep{NEURIPS2019_9015}, coupled with the exponential growth of digital data generation, have collectively steered us into this transformative era of data-centric problem-solving. Today, the profound impact of deep learning on our daily lives is further exemplified by state-of-the-art large language models (LLMs)~\citep{zhao2023survey} and their compelling applications, such as ChatGPT~\citep{ChatGPT}.

\begin{figure}
    \centering
    \includegraphics[width=0.5\linewidth]{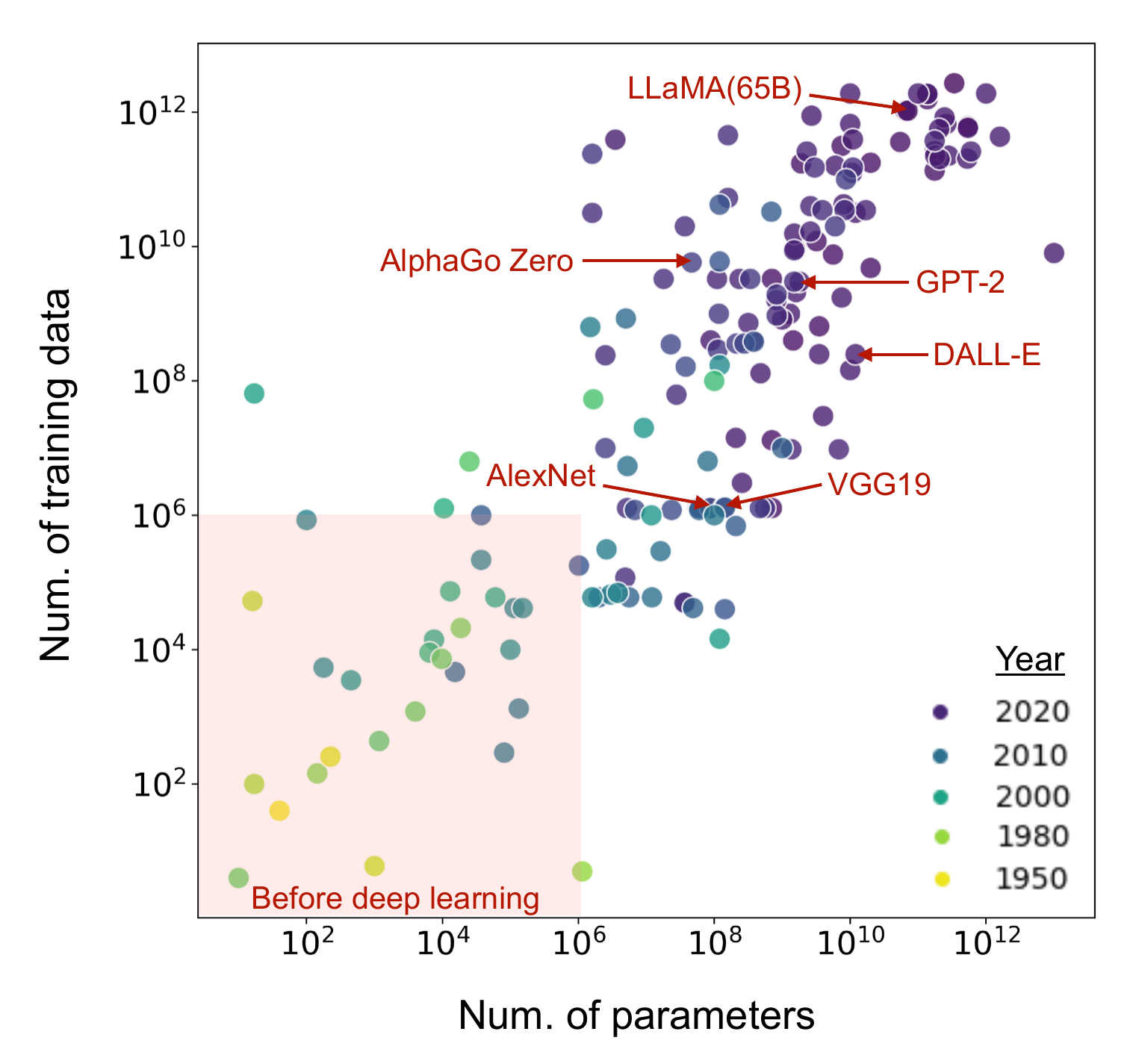}
    \caption{Evolution of machine learning model size from the perspective of number of parameters and number of training data~\citep{Villalobos2022MachineLM}. Some of the famous deep learning models are labeled: AlexNet~\citep{alexnet:2012}, VGG19~\citep{VGG19:2015}, AlphaGo Zero~\citep{Silver:2017aa}, DALL-E~\citep{DALLE:2021}, GPT-2~\citep{GPT2:2019}, LLaMA(65B)~\citep{touvron2023llama}}
    \label{fig:numpara}
\end{figure}

The idea of using data in modeling and problem solving is not new; indeed, this was the foundation upon which empirical science was built. What distinguishes the Fourth Paradigm is the astoundingly high performance of deep learning, the linchpin of modern machine learning, especially when paired with big data. Traditional statistical theories struggled to elucidate the behavior of deep learning models, sparking a drive to devise new statistical theories tailored for contemporary machine learning. While we are yet to arrive at definitive answers, Table~\ref{tab:stattheo} summarizes some preliminary findings from the latest research. Classic statistical models typically employ a restrained number of parameters, prioritizing interpretability and sidestepping overfitting. In contrast, deep learning architectures, particularly deep neural networks, might possess millions or even billions of parameters (Figure~\ref{fig:numpara}). Such magnitude of parameters is traditionally deemed by classical theory to hinder generalizability. Challenging this view, new theories demonstrated that deep architectures can adeptly learn complex, non-smooth functions, outperforming their classic counterparts~\citep{pmlr-v89-imaizumi19a,Imaizumi:2022}. Additionally, the deep frameworks facilitate the automatic extraction of low-dimensional features from high-dimensional inputs~\citep{Nakada:2020}. These insights challenge the traditional idea that equates a higher parameter count with increased model complexity. Emerging theories advocate a perspective where model complexity is contingent upon both the data and the specific learning task~\citep{pmlr-v40-Neyshabur15,NIPS2017_b22b257a,Hu:2021aa}. These new understandings provide important guidelines for the design of machine learning applications across various scientific and engineering domains.

\begin{table}
    \centering
    \begin{tabular}{p{0.10\textwidth}|p{0.08\textwidth}|p{0.10\textwidth}|p{0.14\textwidth}|p{0.12\textwidth}|p{0.14\textwidth}|p{0.14\textwidth}}
         & Num. of parameters & Modeling philosophy & Statistical analysis: overfitting & Statistical analysis: continuity & Statistical analysis: hidden dimension & Statistical analysis: model complexity\\
        \hline
        \hline
        Conventional statistics & $10^1$ to $10^4$ & Sparse modeling, model reduction & Min. num. of parameters; avoid overfitting & Weak to discontinuity in data & Manual feature extraction & Mainly based on number of parameters\\
        \hline
        Modern (big data) statistics & $10^4$ to $10^{12}$ & Deeper layers, over-parameter-ization & More parameters \& deep structure improve accuracy & Efficient to non-linear discontinuity & Extract low dimension features from high dimension input & Based on data dependent model subsets\\
    \end{tabular}
    \caption{Comparison between conventional statistical theories and modern statistical theories.}
    \label{tab:stattheo}
\end{table}

\section{Challenges}

Despite the remarkable advancements brought about by the data-driven approach in science, it is clear there are no easy wins. The deployment of these technologies often demands significant initial investments in data generation, model training hardware, training personnel in data analysis skills, and more. Bridging the gap between informatics and certain traditional experience-driven fields in science and engineering is not straightforward. In this section, we will explore the challenges that hinder the widespread adoption of a data-driven approach in geotechnics. This will be followed by an in-depth discussion on potential opportunities to address these challenges, drawing inspiration from the recent successes of materials informatics, in the subsequent section.

\subsection{Complexity of soil}

\begin{figure}
    \centering
    \includegraphics[width=0.6\linewidth]{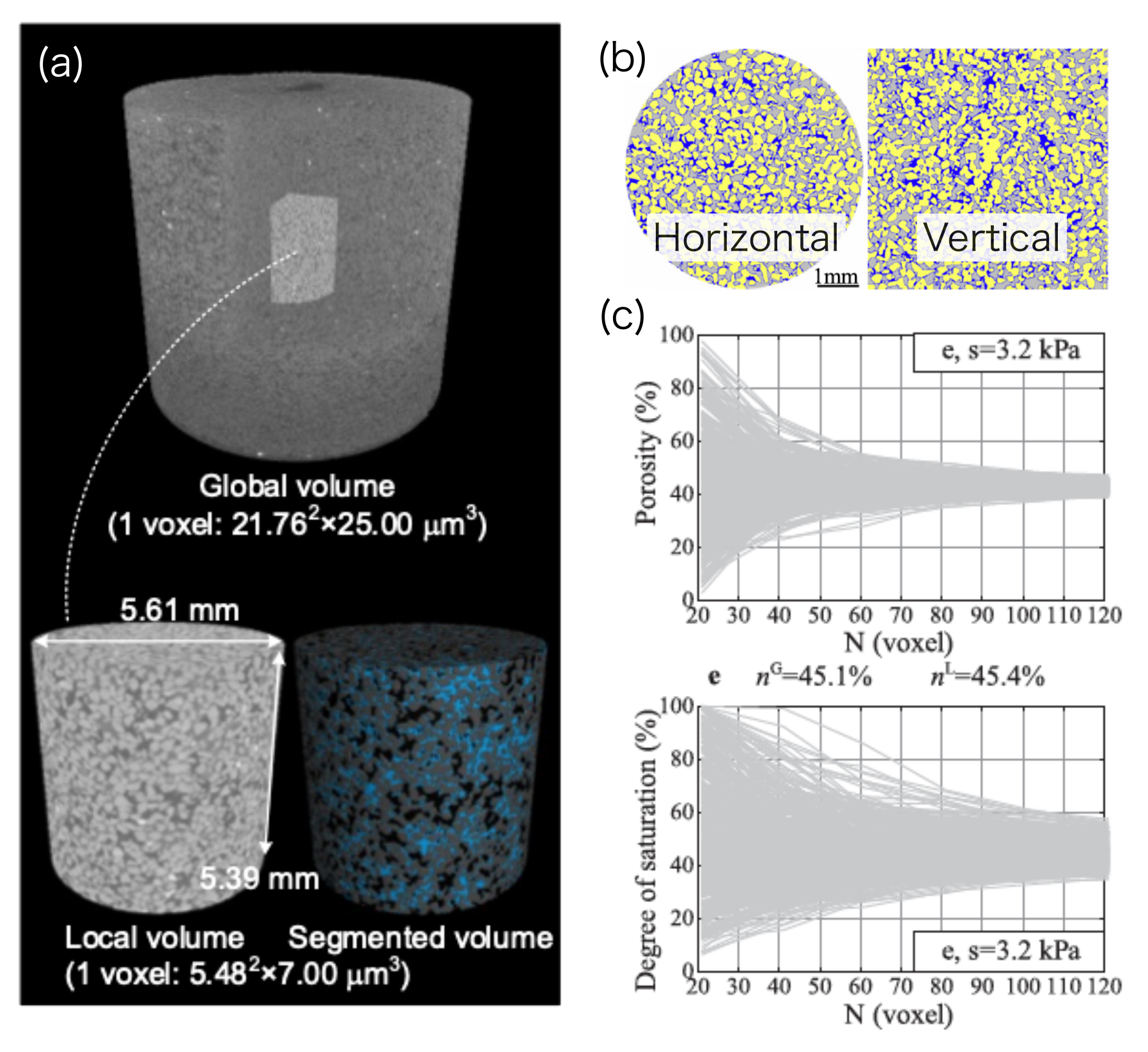}
    \caption{Porosity and saturation variability of unsaturated soil analyzed from multi-scale X-ray Computed Tomography (CT) images. (a) An X-ray CT 3D image of a partially saturated sand specimen and its segmented image~\citep{Higo:2023}. (b) Trinarized images of local CT image (yellow: soil particle, blue: pore water, gray: pore air)~\citep{HIGO20181}. (c) Local quantities of the porosity and degree of saturation with varying subset sizes~\citep{HIGO20181}.}
    \label{fig:soil}
\end{figure}

Soil, at its essence, is an incredibly complex and heterogeneous material, exhibiting properties that vary significantly over multiple scales (Figure~\ref{fig:soil}). At the microscale, individual soil particles, derived from various parent materials, have unique shapes, sizes, and mineralogical compositions. These particles, when bonded together, form intricate pore networks that influence the soil's mechanical and hydraulic behaviors. Scaling up, the mesoscale captures the interactions and arrangements of soil aggregates, or clumps of bonded soil particles. These arrangements are responsible for the macroscopic soil structure, which is crucial in dictating properties such as permeability, shear strength, and compressibility.

Beyond its multi-scale nature, soil is inherently multiphase (Figure~\ref{fig:soil}). It typically consists of solid particles, water, and air. The interactions between these phases, governed by external factors like temperature, moisture content, and loading, can drastically affect soil behavior. For instance, the transition from unsaturated to saturated conditions can lead to significant changes in soil strength and stiffness. Furthermore, certain soils, such as expansive clays, can undergo substantial volume changes with moisture variations, leading to challenges in infrastructure design and maintenance.

Given this inherent complexity, understanding and predicting soil behavior using a data-driven approach is arguably harder than other materials in materials science and engineering. Translating these intricate, multi-scale, and multi-phase characteristics of soil into quantifiable data suitable for machine learning models demands a depth and breadth of data that may be challenging to obtain. This complexity underscores why the data-driven approach, despite its promise, has not been as seamlessly integrated into geotechnics as it has in other domains.

\subsection{Heterogeneity of soil data}
\begin{figure}
    \centering
    \includegraphics[width=0.7\linewidth]{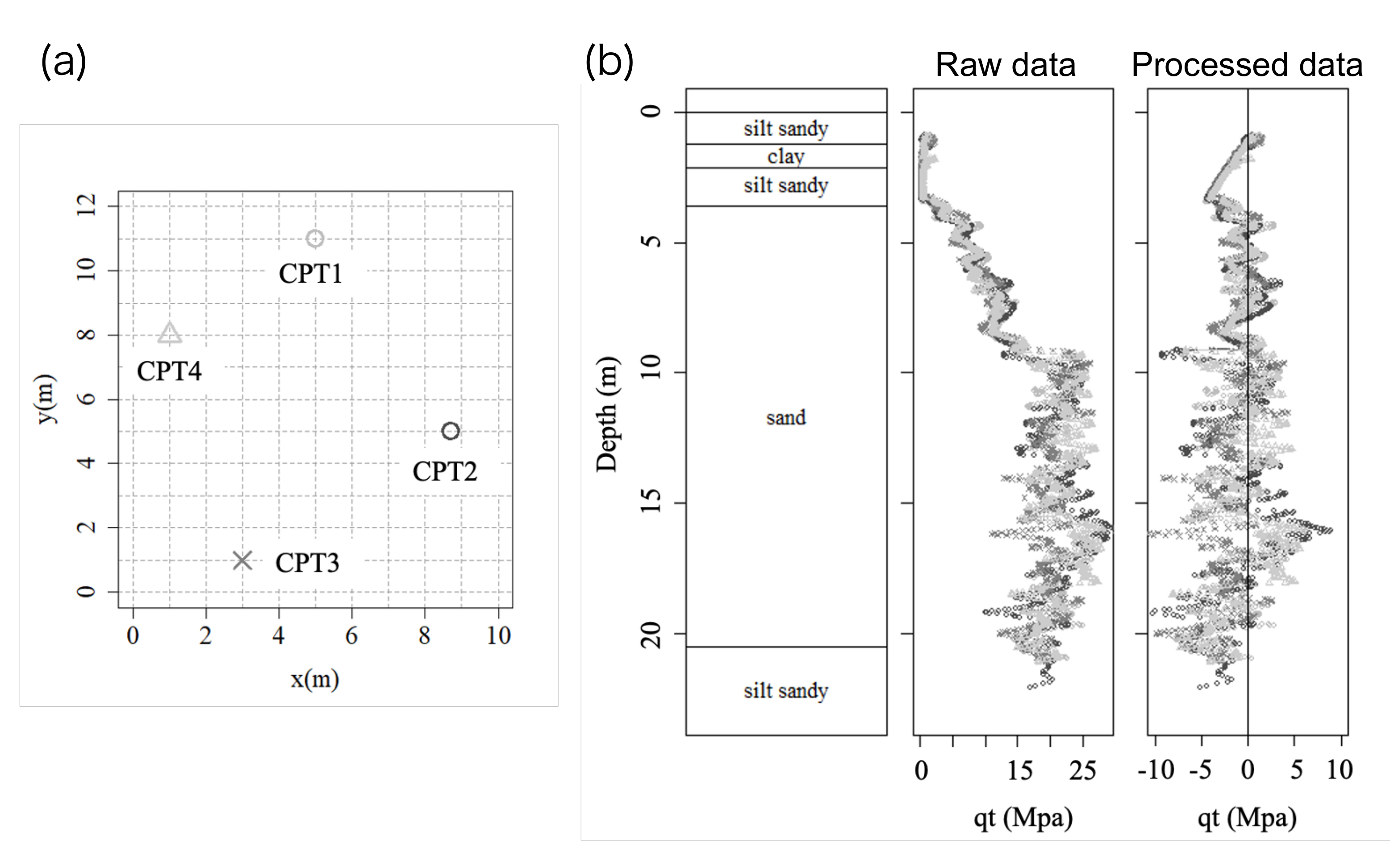}
    \caption{Examples of the heterogeneity of soil data. (a) Plan map showing the locations of Cone Penetration Test (CPT) investigations. (b) Spatial distribution of CPT observation data against depth, comprising both the raw observation data and the processed data with the trend component removed.}
    \label{fig:heter}
\end{figure}

The heterogeneity of soil is another defining characteristics that introduces substantial variability in its properties both spatially and temporally (Figure~\ref{fig:heter}). From one location to another, even over short distances, the soil composition can have surprising variations. This spatial variance in soil properties isn't arbitrary; it's deeply rooted in the geological history of the site. Erosion, sedimentation, tectonic shifts, and volcanic activities, among other processes, have all played roles in shaping the earth's subsurface. Over time, these processes have led to layering, mixing, and reformation of soil, making each site a unique tapestry of geological events. Furthermore, human interventions have introduced another layer of complexity. Urbanization, mining, groundwater extraction, and waste disposal, to name a few, have significantly altered the natural state of soil in many locations. Such anthropogenic activities can lead to changes in soil structure, compaction, contamination, and other modifications, thereby inducing more heterogeneity. Temporal variance adds yet another dimension to the puzzle. Soil properties can change over time due to various factors such as weathering, biological activity, and repeated human activities. This temporal change means that even if soil properties at a specific site are comprehensively understood today, they might evolve and diverge in the future. This creates a challenge, as the behavior of soil at one site might not be representative of its behavior at another, even if they are in close proximity.

This pronounced heterogeneity, rooted in both nature's timeline and human history, complicates the application of a ``one-size-fits-all'' data-driven approach in geotechnics. Unlike other materials, where properties might be consistent and standardized, soil demands a more localized and tailored approach. For geotechnical professionals looking to apply a data-driven approach, the site-dependent nature of soil necessitates the gathering of vast, detailed, and site-specific datasets, presenting logistical and resource challenges. The site characterization challenge has been identified as one of the crucial topics to tackle in order to promote the data-centric geotechnics agenda~\citep{PCS:2022,PHOON2022967}. Recent efforts include using hierarchical Bayesian models~\citep{HBMCLAY:2021} to simultaneously learn the inter- and intra-site variations of soil properties along with clustering algorithms based on prior knowledge~\citep{WU2022102253,SHARMA2023105624}. The inevitable challenge to handle high uncertainties of soil data leads to geotechnics researchers paying a much higher attention to probabilistic learning by Bayesian methods than the latest deep learning methods~\citep{PHOON2022101189}. Based on the list of machine learning research in geotechnics compiled by ISSMGE TC304/TC309 in 2021, 42\% of 444 publications are related to Bayesian methods, while only 27\% are related to artificial neural networks (a combination of deep learning and classical neural networks with a shallow architecture)~\citep{PHOON2022967}.

\subsection{Lack of data}
One of the foundational necessities for the effective application of machine learning and data-driven techniques is the availability of large, diverse and robust datasets. However, in geotechnics, acquiring such datasets is fraught with challenges. On-site experiments --- vital for understanding the behavior and properties of soils in their natural environment --- are not only expensive but also technically difficult to conduct. Given the site-dependent nature of soil properties, obtaining a representative dataset necessitates extensive on-site testing across diverse locations, which is both time-consuming and resource-intensive. Adding to this is the uncertainty introduced during laboratory experiments. Maintaining the exact boundary conditions and natural states of soil samples in a laboratory setup is not a easy task. Any deviation from the natural state can lead to data that, while precise in a lab environment, might not translate perfectly to real-world scenarios.

Yet another significant impediment is the prevalent industry practice of confidentiality. Data acquired from site investigations and experiments, given its cost, sensitivity to owner privacy and strategic importance, is often treated as proprietary information by companies. This proprietary stance acts as a deterrent to open data sharing, resulting in information that remain inaccessible to the broader geotechnical community. The outcome is a fragmented landscape where insights remain locked, which hinders collaborative progress. These challenges further affect the development of a dedicated platform for data sharing. Unlike some other scientific domains where centralized databases and platforms facilitate the dissemination of research data, geotechnics lacks such an infrastructure. This not only limits access to valuable datasets but also hinders the development of standardized data formats and protocols. Some preliminary efforts have been made to alter the situation~\citep{TC304dB}, but we are still far from the scale of data that is typically used in other informatics domains driven by deep learning.

Some researchers describe the geotechnics data as ``ugly''~\citep{PCS:2022}, referring to the Multivariate, Uncertain and unique, Sparse, Incomplete, and potentially Corrupted (MUSIC) features of the data~\citep{BID:2019}. This concept is further extended to MUSIC-3X~\citep{PCS:2022}, covering the three dimensional spatial variations of the data on top of the MUSIC features caused by the complexity and heterogeneity of soil data. In theory, there has been vast amount of geotechnics data accumulated in the long history of the domain. In that sense, the actual bottleneck to the success of GtI is the lack of ``useful'' data. In other words, the existing big data in geotechnics is ``Big Indirect Data'' that is not compatible with modern machine learning framework~\citep{BID:2019}.

% \subsection{Restricted modeling capacity using low dimensional input}

\section{Opportunities}

\begin{figure}
    \centering
    \includegraphics[width=0.8\linewidth]{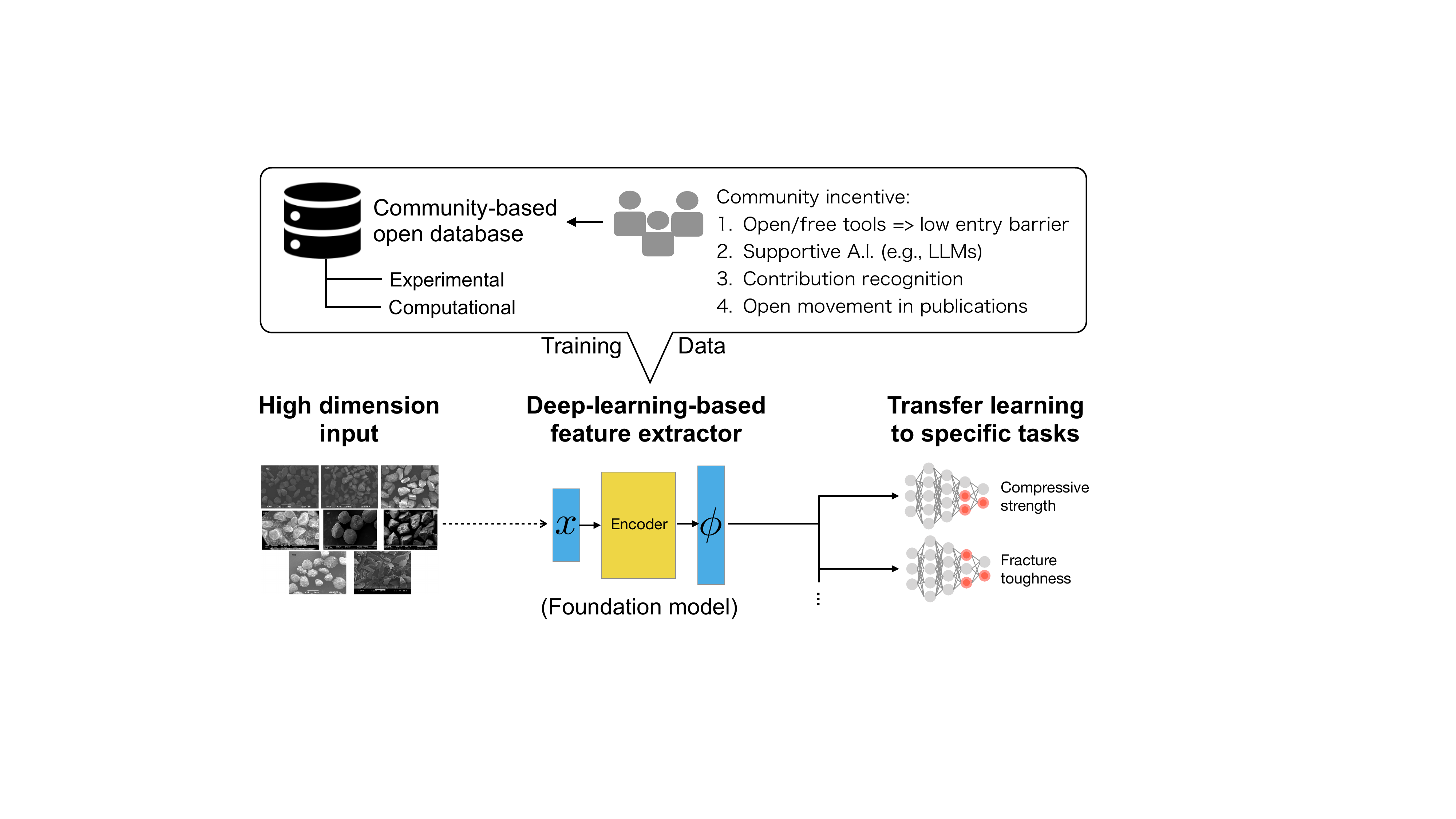}
    \caption{Proposed pathway to a fully data-driven geotechnics.}
    \label{fig:goal}
\end{figure}

The overview in the previous section sheds light on the richness and intricacies of soils, emphasizing the challenges faced by researchers aiming to leverage modern data-driven techniques in geotechnics. Capturing the breadth of variability in soil properties by machine learning models may seem to be a formidable task. However, with the right tools and methodologies, these challenges can be transformed into opportunities for groundbreaking advancements in the field of GtI. Figure~\ref{fig:goal} summarizes the opportunities discussed in this section.

\subsection{New problem setup in high dimensional space}

One of the key factors contributing to the success of deep learning is its ability to automatically extract useful features from vast amounts of data residing in high-dimensional spaces. For instance, the convolutional layer in CNNs is a generalization of image filters, proven highly effective in extracting representative patterns from natural images for image recognition tasks~\citep{Alzubaidi:2021aa}. The Transformer attention mechanism stands at the heart of the remarkable success in understanding the sequential features of natural languages~\citep{NIPS2017_3f5ee243}. Similarly, the Fourier Neural Operator has showcased an impressive capacity to capture essential periodic patterns, thereby enabling accurate predictions for a range of complex partial differential equations~\citep{li2021fourier}. When paired with the appropriate dataset, deep learning continues to underscore its prowess in capturing intricate features, often surpassing human imagination.

In the field of geotechnical engineering, granular materials are traditionally represented by modeling them as a continuous body. Over the years, geotechnics experts have identified and established various empirical parameters that correlate with distinctive soil properties of practical significance, which include grain-size distribution curves, void ratios, and moisture content ratios. These parameters are often referred to as indicative properties or indicative parameters, and they are anticipated to play a pivotal role in conveying the authentic behavior of granular soil materials in mechanical computations. Nevertheless, their utilization in practical applications remains insufficient.

An alternative method for estimating soil's mechanical properties is by analyzing data from standard and cone penetration tests (SPT and CPT) conducted at the site. These tests, especially the CPT which offers continuous soil property assessment with depth, provide key correlations used in empirical equations like the Khowy Mahye for mechanical estimation. While these empirical methods, backed by substantial data and acknowledged safety margins, have been crucial in structural design, they are also known to carry inherent uncertainties that represent a significant source of risk in geotechnical design, as noted by~\citet{OTAKE2022101129}. Current efforts to integrate machine learning with geotechnics predominantly hinge on these parameters or low dimension vectors obtained from classic dimension reduction algorithms. Typically, the dimensionality of these parameters does not exceed a few tens, with most studies working with less than ten dimensions~\citep{GS:2019,HBMCLAY:2021,LI2022106769}. Given the noisy and incomplete character of geotechnical data, as introduced in previous sections, many studies often resort to utilizing only a subset of the complete range of indicator parameters for specific predictive tasks~\citep{WU2022102253,LI2022106769,OZSAGIR2022105014}. This approach, however, poses limitations: regardless of the volume of training data available, it remains a challenge to encapsulate the extensive variability of soil properties through such machine learning models.

\begin{figure}
    \centering
    \includegraphics[width=0.9\linewidth]{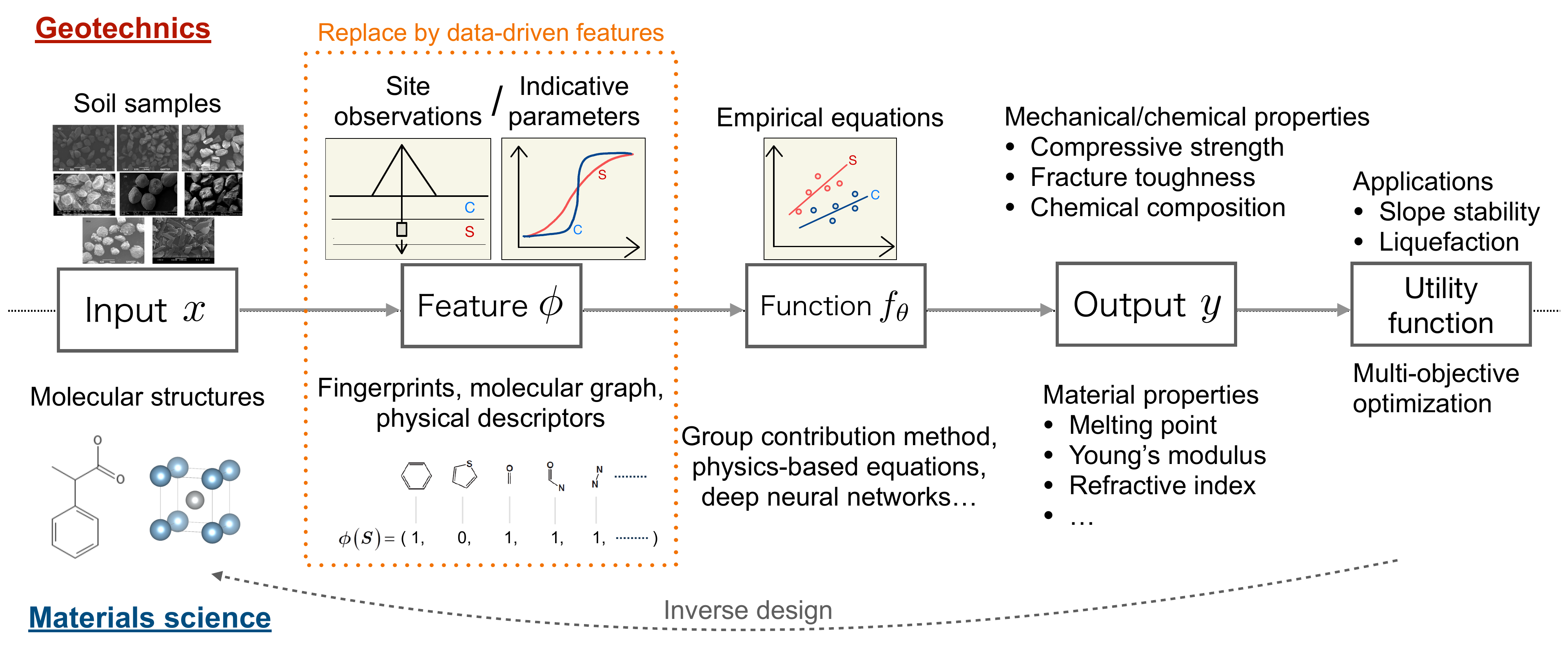}
    \caption{Typical problem setup of geotechnics and materials science.}
    \label{fig:problem}
\end{figure}

Materials science shares a similar problem setup to geotechnics (Figure~\ref{fig:problem}). In light of the early successful examples of implementing a data-driven approach to materials design~\citep{MK:2016,Tabor:2018,Wu:2019aa}, many researchers began to replace the conventional descriptors designed based on human insights by artificial descriptors extracted by unsupervised learning~\citep{GB:2018,Gupta:2022aa} or automated algorithms~\citep{ECFP:2010}. The machine-learned descriptors have played an important role in the rapid development of MI, proving their ability to uncover lost information when reducing high dimension data to low dimension empirical parameters by human instinct. Learning of such descriptors often only requires unlabeled materials data, such as 1D, 2D, or 3D molecular representations, to be trained with unsupervised learning algorithms. This approach is the next essential step to take for triggering a paradigm shift in geotechnics. 

\subsection{New database design}
\begin{figure}
    \centering
    \includegraphics[width=0.7\linewidth]{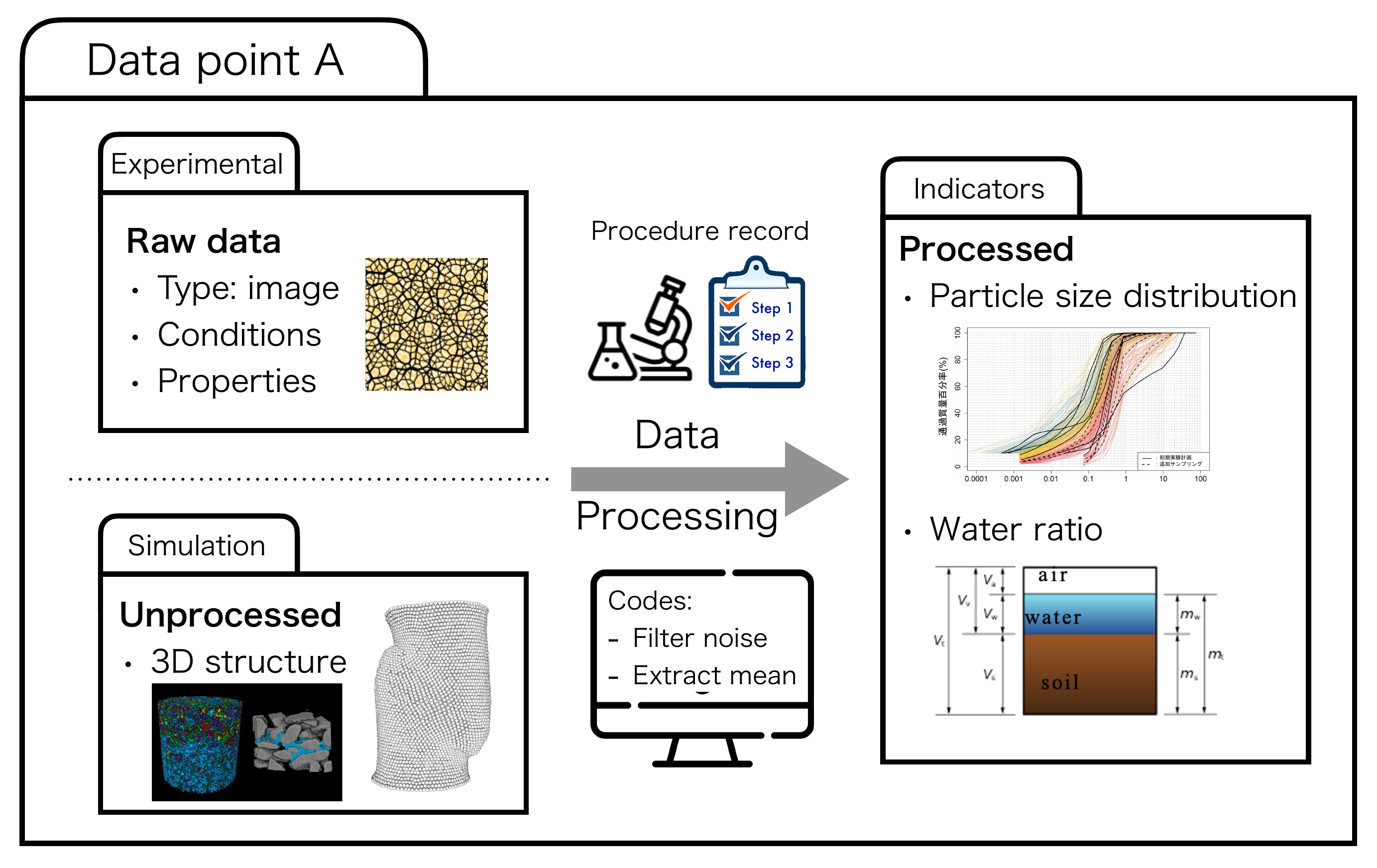}
    \caption{Design of a new database for GtI. Each data point can contain experimental and/or computational data. Raw data is always stored along with the processed data. The processing procedures are also stored to ensure reproducibility.}
    \label{fig:database}
\end{figure}

Modern machine learning frameworks, particularly those based on deep neural network architectures, exhibit a remarkable flexibility in handling a wide array of input data types. This encompasses everything from images, videos, and natural language to continuous spectral data. This versatility, combined with the prowess of deep learning's feature extraction capabilities, has significantly expanded the horizons of what types of data can be used for training models. Consequently, this expansion has sparked new design philosophies for constructing databases tailored to support these advanced modeling methods. Figure~\ref{fig:database} shows one of the possible design concepts.

To fully harness the capability of deep learning, it is imperative to store raw, high-dimensional data before it undergoes any processing into a lower dimensionality. This approach facilitates the use of multiple small datasets which can be leveraged using methods like multitask learning and transfer learning, both of which show promise in effectively utilizing sparse or limited data. Finding the right equilibrium between the size of the data and the depth of information each data point provides becomes increasingly difficult, prompting a shift towards databases designed with a hierarchical structure. Within such a structure, every data processing step is documented in detail, ensuring clarity and reproducibility. An ideal database would incorporate data processing methods which are algorithm-driven, ensuring consistency and reliability. These algorithms would be stored within the database itself for future reference and application. Another vital aspect to consider is the explicit recording of uncertainties associated with the data, given the inherent variability and challenges of geotechnical measurements.

In situations where experimental data might be scarce, there is an emerging consensus about pivoting towards computational data derived from physics-based simulations. MI provides a compelling example in this realm. Many of MI's early achievements stemmed from databases that relied heavily on first-principle calculations via simulations~\citep{materialsproject,Kirklin:2015aa,ramakrishnan2014quantum}. Another intriguing advancement in the MI sphere is the use of robotic systems for automated generation of experimental data~\citep{Burger:2020aa,MacLeod2020,Vriza:2023aa}, a method that is at the forefront of database enrichment.

Crafting modern databases that align with the demands and intricacies of cutting-edge machine learning is no solitary venture. It requires collaborative efforts from the community, pooling expertise, resources, and insights to build a cohesive, robust, and adaptive information repository.

\subsection{Transfer learning technology to bridge the gaps}
The marvel of human learning lies in our innate ability to quickly master new tasks with minimal exposure. Given only a few examples, humans can proficiently deduce and apply knowledge, a skill often facilitated by extrapolating from a vast array of prior experiences across different domains. Inspired by this intrinsic human capability, the concept of transfer learning emerged in the realm of machine learning. Transfer learning is essentially about leveraging the knowledge extracted from one task (the source) to improve the learning of another, often related, task (the target) with limited data.

The rise of deep learning has brought transfer learning to the forefront, with its principles becoming almost ubiquitous in contemporary machine learning practices. The essence of transfer learning resonates with the architectural flexibility and adaptability of deep neural networks, leading to the development of a rich set of strategies to implement it effectively~\citep{Weiss:2016aa,TLsurvey:2018}. One domain where the impact of transfer learning is profoundly felt is materials informatics. In the context of materials science, there's often a lack of data for many problems of practical significance. Transfer learning emerges as a linchpin in such scenarios, enabling researchers to leverage data from related tasks to bridge the data gap~\citep{Yamada:2019aa}.

A fascinating application of transfer learning is its ability to act as a bridge between simulation data and real-world data. Simulations, being based on theoretical models, often provide idealized outcomes, while real-world data captures the nuances and unpredictabilities of practical environments. Transfer learning aids in harmonizing these two realms, ensuring that knowledge from simulations can be adapted effectively to real-world scenarios. Additionally, transfer learning also provides an avenue to harmonize large data domains with smaller ones. In instances where abundant data might be available for a particular aspect but is scarce for a closely related domain, transfer learning can be the bridge that ensures knowledge flow between them~\citep{WANG2022}.

Extending this principle further, the geotechnics domain can significantly benefit from transfer learning by adapting insights from other fields~\citep{Kim2018TL,ICIIP2021TL,Eyob:2024aa}. Given the intricate, heterogeneous nature of soil and its interactions, geotechnics can draw upon data from related scientific domains, ensuring a more holistic understanding and application. In essence, transfer learning in GtI paves the way for a more interconnected and comprehensive approach to studying and addressing the challenges the field presents.

\subsection{Open science movement}

The advancement of GtI stands at a pivotal intersection where the spirit of collaboration, transparency, and accessibility offered by the open science movement plays a crucial role~\citep{Woelfle:2011aa}. As we tread into an era characterized by the convergence of traditional disciplines and advanced computational techniques, the importance of democratizing knowledge and tools becomes more evident than ever. The open science movement has fundamentally revolutionized the landscape of scientific research, fostering a collaborative culture and democratizing access to knowledge. One of the most profound impacts of this movement has been seen in the domain of machine learning. Open source tools for machine learning and data analysis have significantly leveled the playing field, enabling even those without a deep expertise in data science to contribute to, and benefit from, the vast strides the field has made. Platforms like TensorFlow, PyTorch, and Scikit-learn~\citep{scikit-learn}, among others, have ensured that the power of modern computation is not confined to the domain of data scientists alone. In MI, RDKit~\citep{RDKit} has become the fundamental digital toolbox for analyzing organic materials. By removing barriers to entry, these open source tools have played an indispensable role in the ubiquity and success of machine learning applications across various sectors.

Equally transformative has been the growing emphasis on open data and code in machine learning publications. By mandating the accessibility of raw data and the associated computational scripts, the scientific community ensures that the research is transparent, reproducible, and lends itself to further enhancements by peers. This culture of openness resonates with the FAIR data movement's principles, which advocates for scientific data to be Findable, Accessible, Interoperable, and Reusable~\citep{Wilkinson:2016aa}. Such principles, when embedded within the geotechnics community, can dramatically elevate the quality and impact of research. Despite its potential benefits, open data and code practices are not yet pervasive in geotechnics publications. There is a compelling need to usher in a change, especially considering the intricacies and heterogeneity characteristic of geotechnical data. 

Embracing an open science framework extends beyond just sharing; it actively adds value to the intricate process of data generation in geotechnics. This is where the idea of community-driven databases becomes attractive. By fostering community-driven databases, every piece of data, every experiment, and every computational model becomes part of a larger tapestry of shared knowledge, enhancing its significance multifold. Such databases become living entities, continually evolving and expanding with collective contributions. It is important to note that the act of sharing should not be perceived as a mere act of charity. The open science movement recognizes the efforts and contributions of individuals and institutions. By streamlining the data sharing process and offering incentives, be it in the form of recognition, easier publication pathways, or even tangible benefits, the movement ensures that participants are motivated and find value in their act of sharing. 

In certain situations, conflicts of interest can emerge due to industry competition. This has been a significant obstacle in gathering information from industrial players who possess extensive data in MI. However, a recent initiative named RadonPy~\citep{Hayashi:2022aa} successfully overcame this challenge by uniting nearly all major chemical companies in Japan to develop the world's first fully open database for polymeric materials. By committing to lay a foundation for propelling the field forward, participants can foresee new opportunities that align with their own interests. Employing a similar strategy might be effective in encouraging the geotechnics community to harness the collective intelligence and insights of its members, thus fueling innovation and propelling the discipline to new heights.

%\subsection{Embracing new AI tools}
%- LLM in MI as an example

\section{Conclusion}
In the rapidly evolving landscape of modern computational science, the geotechnics domain stands at a critical juncture. Historically grounded in empirical and often experience-driven methodologies, geotechnics has accumulated vast amounts of data over time. This data, rooted in the complex, multi-scale, multi-phase, and inherently heterogeneous nature of soil, presents both a challenge and an opportunity for integration with advanced computational models. Challenges arise from the intricacy of soil properties, the significant spatial and temporal variances due to geological history, and the often-limited availability of data. The latter is a consequence of the high costs associated with on-site experiments, uncertainties in lab-based experiments, the prevailing culture of confidentiality in industry, and the absence of robust data-sharing platforms.

Yet, the dawn of GtI brings forth a renewed hope. The exceptional prowess of deep learning to autonomously distill meaningful features from high-dimensional data has already marked its success in various fields. This characteristic, combined with innovative database designs tailored for deep learning, holds immense potential. By adopting techniques like transfer learning, we can effectively harness data from diverse sources, bridging gaps between simulation and reality, between vast and limited data realms, and even integrating knowledge from disparate fields. The path forward is further illuminated by the open science movement. The commitment to open-source tools, adherence to the FAIR data principles, and fostering community-driven data-sharing platforms could catalyze unprecedented advancements in geotechnics.

In conclusion, while GtI confronts certain hurdles, the infusion of modern machine learning technologies, especially deep learning, offers a promising avenue. Embracing collaborative efforts, as seen in the success of MI, could pave the way for harnessing collective expertise. As the geotechnics community approaches this new horizon, the advent of sophisticated computational tools like LLMs, including ChatGPT, further augments this promise. These models not only provide advanced data analytics but also enable intuitive human-computer interactions, making the interpretation and utilization of intricate geotechnical data more accessible to both experts and novices alike. Such expectation is well supported by the many successful examples of exploiting LLMs in MI~\citep{Zheng:2023aa,D3DD00113J}. Looking forward, we anticipate a dynamic confluence of traditional geotechnical knowledge with these state-of-the-art computational methodologies. This fusion promises not just a deeper understanding but also paves the way for innovative solutions in geotechnics. 

% \appendix
% \section{My Appendix}
% Appendix sections are coded under \verb+\appendix+.

% \verb+\printcredits+ command is used after appendix sections to list 
% author credit taxonomy contribution roles tagged using \verb+\credit+ 
% in frontmatter.

\section{Acknowledgments}
This research was supported by ``Strategic Research Projects'' grant from ROIS (Research Organization of Information and Systems). This paper was proofread using ChatGPT-4 on October 30, 2023, for correcting grammatical mistakes.

\printcredits

\bibliographystyle{cas-model2-names}

\bibliography{GtI_ref_list}


% Author biography without author photo.
% Author biography. Author biography. Author biography.
% Author biography. Author biography. Author biography.
% Author biography. Author biography. Author biography.
% Author biography. Author biography. Author biography.
% Author biography. Author biography. Author biography.
% Author biography. Author biography. Author biography.
% Author biography. Author biography. Author biography.
% Author biography. Author biography. Author biography.
% Author biography. Author biography. Author biography.

% \bio{figs/pic1}
% Author biography with author photo.
% Author biography. Author biography. Author biography.
% Author biography. Author biography. Author biography.
% Author biography. Author biography. Author biography.
% Author biography. Author biography. Author biography.
% Author biography. Author biography. Author biography.
% Author biography. Author biography. Author biography.
% Author biography. Author biography. Author biography.
% Author biography. Author biography. Author biography.
% Author biography. Author biography. Author biography.

% \bio{figs/pic1}
% Author biography with author photo.
% Author biography. Author biography. Author biography.
% Author biography. Author biography. Author biography.
% Author biography. Author biography. Author biography.
% Author biography. Author biography. Author biography.

\end{document}